\documentclass[11pt,a4paper]{article}
\usepackage[hyperref]{eacl2021}
\usepackage{times}
\usepackage{latexsym}

\usepackage{microtype}

\aclfinalcopy % Uncomment this line for the final submission
\def\aclpaperid{903} %  Enter the acl Paper ID here

\usepackage{amsmath}
\usepackage{hyperref}
\usepackage{url}
\usepackage{mathrsfs}
\usepackage{multirow}
\usepackage{booktabs}
\usepackage{graphicx}
\usepackage{subcaption}
\usepackage{cleveref}
\usepackage{CJK}
\usepackage{CJKspace}
\usepackage{fdsymbol}
\usepackage{enumitem} 

\crefname{equation}{Eq.}{Eq.}
\crefname{section}{Section}{Sections}
\crefname{subsection}{Section}{Sections}
\crefname{subsubsection}{Section}{Sections}
\crefname{figure}{Figure}{Figures}
\crefname{table}{Table}{Tables}
\crefname{subfigure}{Figure}{Figures}
\crefname{algocf}{Algorithm}{Algorithms}
\usepackage{pifont}
\title{\textsc{L2C}: Describing Visual Differences\\ Needs Semantic Understanding of Individuals}

\author{An Yan$^{\vardiamondsuit}$, Xin Eric Wang
$^{\spadesuit}$, Tsu-Jui Fu$^\clubsuit$, William Yang Wang$^\clubsuit$\\
$^\vardiamondsuit$UC San Diego, $^\spadesuit$UC Santa Cruz, $^\clubsuit$UC Santa Barbara\\
\texttt{\small ayan@ucsd.edu, xwang366@ucsc.edu, \{tsu-juifu,william\}@cs.ucsb.edu}\\}

\date{}

\begin{document}
\maketitle

\begin{abstract}
% no more than 200 words
Recent advances in language and vision push forward the research of captioning a single image to describing visual differences between image pairs. 
Suppose there are two images, $I_1$ and $I_2$, and the task is to generate a description $W_{1,2}$ comparing them, existing methods directly model $\langle I_1, I_2 \rangle \rightarrow W_{1,2}$ mapping without the semantic understanding of individuals. 
In this paper, we introduce a Learning-to-Compare (\textsc{L2C}) model, which learns to understand the semantic structures of these two images and compare them while learning to describe each one.   
We demonstrate that \textsc{L2C} benefits from a comparison between explicit semantic representations and single-image captions, and generalizes better on the new testing image pairs. 
It outperforms the baseline on both automatic evaluation and human evaluation 
for the Birds-to-Words dataset.
\end{abstract}

\section{Introduction}
\label{sec:intro}
The task of generating textual descriptions of images tests a machine's ability to understand visual data and interpret it in natural language. 
It is a fundamental research problem lying at the intersection of natural language processing, computer vision, and cognitive science.
For example, single-image captioning~\citep{farhadi2010every, kulkarni2013babytalk, vinyals2015show, xu2015show} has been extensively studied.

\begin{figure}[htbp]
  \centering
  \includegraphics[width=.9\linewidth]{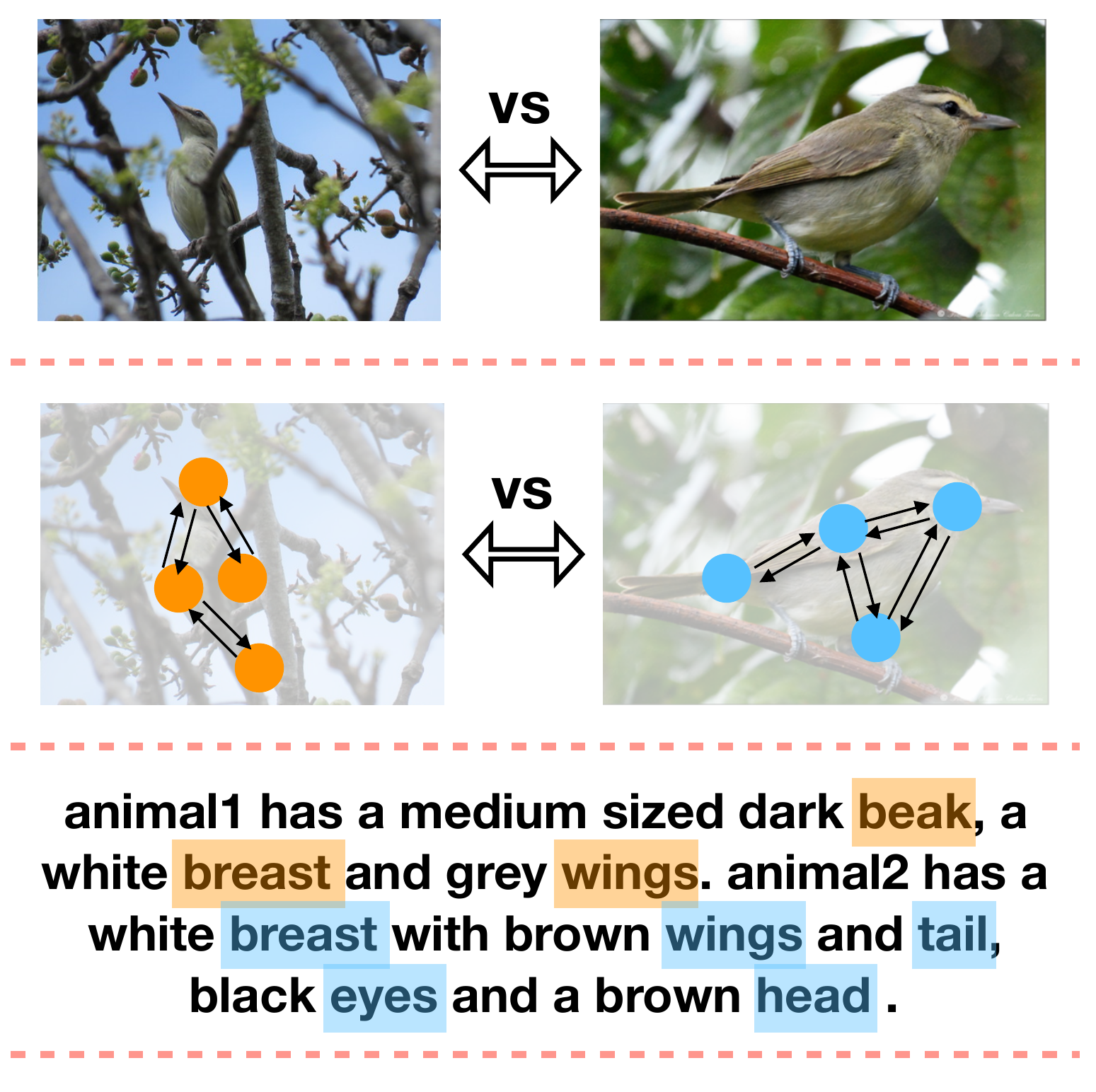}
  \caption{Overview of the visual comparison task and our motivation. The key is to understand both images and compare them. Explicit semantic structures can be compared between images and used to generate comparative descriptions aligned to the image saliency.}
  \label{fig:task}
\end{figure}

Recently, a new intriguing task, visual comparison, along with several benchmarks ~\citep{jhamtani2018learning, tan2019expressing, park2019robust, forbes2019neural} has drawn increasing attention in the community.
To complete the task and generate comparative descriptions, a machine should understand the visual differences between a pair of images (see \cref{fig:task}).
Previous methods~\cite{jhamtani2018learning} often consider the pair of pre-trained visual features such as the ResNet features~\cite{he2016deep} as a whole, and build end-to-end neural networks to predict the description of visual comparison directly.
In contrast, humans can easily reason about the visual components of a single image and describe the visual differences between two images based on their semantic understanding of each one. 
Humans do not need to look at thousands of image pairs to describe the difference of new image pairs, as they can leverage their understanding of single images for visual comparison. 

\begin{figure*}[t]
% \vspace{-2ex}
\centering
\includegraphics[width=\textwidth]{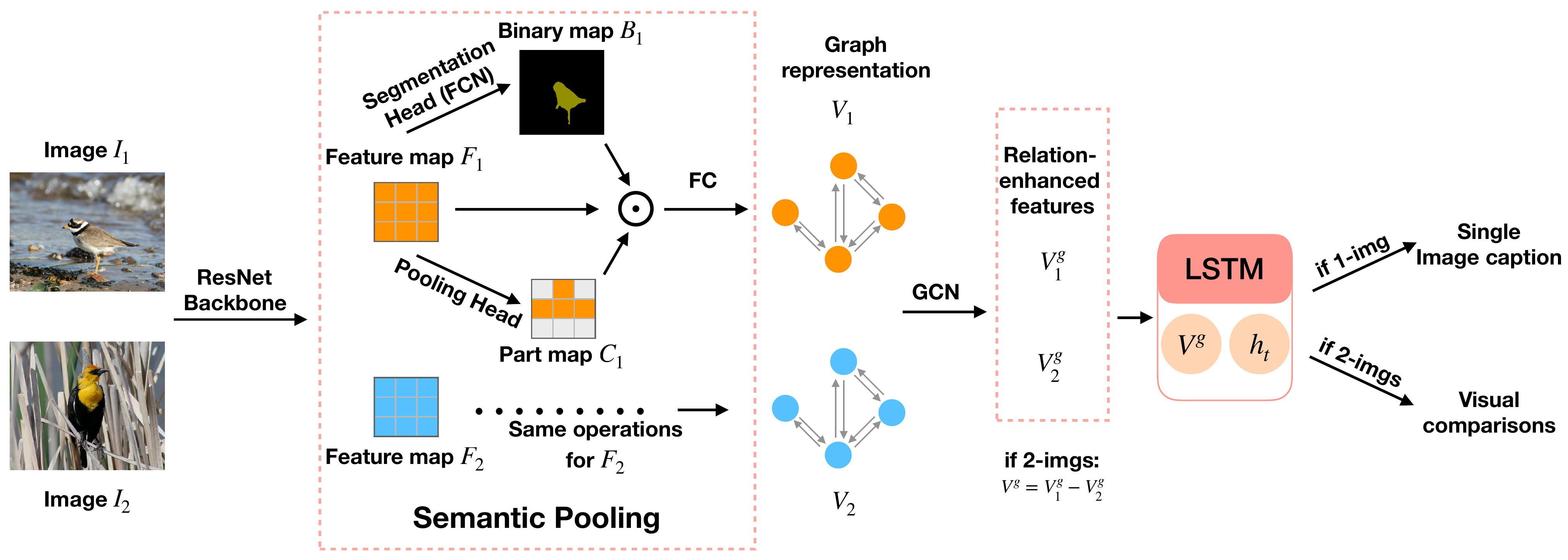}
\caption{Our \textsc{L2C} model. It consists of a segmentation encoder, a graph convolutional module, and an LSTM decoder with an auxiliary loss for single-image captioning. Details are in \cref{sec:method}.}
\label{fig:model}
\end{figure*}

Therefore, we believe that visual differences should be learned by understanding and comparing every single image's semantic representation.
A most recent work~\cite{zhang2020diagnosing} conceptually supports this argument, where they show that low-level ResNet visual features lead to poor generalization in vision-and-language navigation, and high-level semantic segmentation helps the agent generalize to unseen scenarios. 

Motivated by humans, we propose a Learning-to-Compare (\textsc{L2C}) method that focuses on reasoning about the semantic structures of individual images and then compares the difference of the image pair. 
Our contributions are three-fold: 
\begin{itemize}
    \setlength\itemsep{-0.2em}
    \item We construct a structured image representation by leveraging image segmentation with a novel semantic pooling, and use graph convolutional networks to perform reasoning on these learned representations.
    \item We utilize single-image captioning data to boost semantic understanding of each image with its language counterpart.
    \item Our \textsc{L2C} model outperforms the baseline on both automatic evaluation and human evaluation, and generalizes better on the testing image pairs.
\end{itemize}

\section{\textsc{L2C} Model}
\label{sec:method}
We present a novel framework in \cref{fig:model}, which consists of three main components. 
First, a \emph{segmentation encoder} is used to extract structured visual features with strong semantic priors.
Then, a \emph{graph convolutional module} performs reasoning on the learned semantic representations. 
To enhance the understanding of each image, we introduce a \emph{single-image captioning auxiliary loss} to associate the single-image graph representation with the semantic meaning conveyed by its language counterpart.
Finally, a decoder generates the visual descriptions comparing two images based on differences in graph representations. 
All parameters are shared for both images and both tasks.

% \subsection{Constructing Semantic Representation}
\subsection{Semantic Representation Construction}
To extract semantic visual features, we utilize pre-trained fully convolutional networks (FCN)~\citep{long2015fully} with ResNet-101 as the backbone. 
An image $\mathcal{I}$ is fed into the ResNet backbone to produce a feature map $\mathcal{F}$ $\in \mathbb{R}^{D\times H\times W}$, which is then forwarded into an FCN head that generates a binary segmentation mask $B$ for the bird class. 
However, the shapes of these masks are variable for each image, and simple pooling methods such as average pooling and max pooling would lose some information of spatial relations within the mask.
% lose some spatial relationship within the mask to some extent. 

To address this issue and enable efficient aggregation over the area of interest (the masked area), we add a module after the ResNet to cluster each pixel within the mask into $K$ classes. Feature map $\mathcal{F}$ is forwarded through this pooling module to obtain a confidence map $\mathcal{C}$ $\in \mathbb{R}^{K\times H\times W}$, whose entry at each pixel is a $K$-dimensional vector that represents the probability distribution of $K$ classes.

Then a set of nodes $V = \{v_1, ..., v_K\}, v_k \in \mathbb{R}^D$ is constructed as following: 
\begin{equation}
    v_k= \sum_{i, j} \mathcal{F} \odot \mathcal{B} \odot \mathcal{C}_k
\end{equation}
where $i$=$1,... H,$ $j$=$1,...,W ,$, $\mathcal{C}_k$ is the $k$-th probability map and $\odot$ denotes element-wise multiplication.

To enforce local smoothness, i.e., pixels in a neighborhood are more likely belong to one class, we employ total variation norm as a regularization term:
\begin{equation}
    \mathcal{L}_{TV} = \sum_{i,j}|C_{i+1,j}-C{i,j}|+|C_{i,j+1}-C{i,j}|
\end{equation}

% \subsection{Visual Relation Reasoning}
\subsection{Comparative Relational Reasoning}
Inspired by recent advances in visual reasoning and graph neural networks ~\citep{chen2018iterative, li2019visual}, we introduce a relational reasoning module to enhance the semantic representation of each image.
% Specifically, we measure the pairwise affinity between the nodes of the semantic graph in the constructed embedding space. 
A fully-connected visual semantic graph $G = (V, E)$ is built, where $V$ is the set of nodes, each containing a regional feature, and $E$ is constructed by measuring the pairwise affinity between each two nodes $v_i, v_j$ in a latent space.
\begin{equation}
    A(v_i, v_j) = (W_i v_i)^T (W_j v_j)
\end{equation}
where $W_i, W_j$ are learnable matrices, and $A$ is the constructed adjacency matrix. 

We apply Graph Convolutional Networks (GCN) ~\citep{kipf2016semi} to perform reasoning on the graph.
After the GCN module, the output $V^o = \{v_1^o, ..., v_K^o\}, v_k^o \in \mathbb{R}^D$ will be a relationship enhanced representation of a bird.
For the visual comparison task, we compute the difference of each two visual nodes from two sets, denoted as  $V^g_{diff} = \{v_{diff,1}^o, ..., v_{diff,K}^o\}, v_{diff,k}^o = v_{k,1}^o - v_{k, 2}^o \in \mathbb{R}^D$.

% \subsection{Generating Descriptions}
\subsection{Learning to Compare while Learning to Describe}
After obtaining relation-enhanced semantic features, we use a Long Short-Term Memory (LSTM) ~\citep{hochreiter1997long} to generate captions. 
As discussed in \cref{sec:intro}, semantic understanding of each image is key to solve the task. However, there is no single dataset that contains both visual comparison and single-image annotations.
Hence, we leverage two datasets from similar domains to facilitate training. One is for visual comparison, and the other is for single-image captioning. Alternate training is utilized such that for each iteration, two mini-batches of images from both datasets are sampled independently and fed into the encoder to obtain visual representations $V^o$ (for single-image captioning) or $V^o_{diff}$ (for visual comparison).

The LSTM takes $V^o$ or $V^o_{diff}$ with previous output word embedding $y_{t-1}$ as input, updates the hidden state from $h_{t-1}$ to $h_t$, and predicts the word for the next time step.
The generation process of bi-image comparison is learned by maximizing the log-likelihood of the predicted output sentence. The loss function is defined as follows:
% \vspace{-1ex}
\begin{equation}
    \mathcal{L}_{diff}=-\sum_t {\log P(y_{t}|y_{1:t-1}, V^o_{diff})}
\end{equation}
Similar loss is applied for learning single-image captioning:
\begin{equation}
    \mathcal{L}_{single}=-\sum_t {\log P(y_{t}|y_{1:t-1}, V^o)}
\end{equation}

Overall, the model is optimized with a mixture of cross-entropy losses and total variation loss:
% \vspace{-1ex}
\begin{equation}
    \begin{split}
    \mathcal{L}_{loss} = \mathcal{L}_{diff} + \mathcal{L}_{single} +  \lambda \mathcal{L}_{TV}
    \end{split}
\end{equation}
where $\lambda$ is an adaptive factor that weighs the total variation loss.

\section{Experiments}
\subsection{Experimental Setup}
\paragraph{Datasets} 
The Birds-to-Words (B2W) has 3347 image pairs, and each has around 5 descriptions of visual difference. This leads to 12890/1556/1604 captions for train/val/test splits. Since B2W contains only visual comparisons, We use the CUB-200-2011 dataset (CUB) ~\citep{wah2011caltech}, which consists of single-image captions as an auxiliary to facilitate the training of semantic understanding. 
CUB has 8855/2933 images of birds for train/val splits, and each image has 10 captions.

\paragraph{Evaluation Metrics}
Performances are first evaluated on three automatic metrics\footnote{\url{https://www.nltk.org}}: BLEU-4~\citep{papineni2002bleu}, ROUGE-L~\citep{lin-2004-rouge}, and CIDEr-D~\citep{vedantam2015cider}. Each generated description is compared to all five reference paragraphs. Note for this particular task, researchers observe that CIDEr-D is susceptible to common patterns in the data (See \cref{tab:main} for proof), and ROUGE-L is anecdotally correlated with higher-quality descriptions (which is noted in previous work~\citep{forbes2019neural}). Hence we consider ROUGE-L as the major metric for evaluating performances.
We then perform a human evaluation to further verify the performance.

\begin{table*}[t]
\small
\centering
\setlength{\tabcolsep}{8pt}
\begin{tabular}{l rrrrr rrrrr}
\toprule
 & \multicolumn{3}{c}{\textbf{Validation}} & \multicolumn{3}{c}{\textbf{Test}}\\
\cmidrule(lr){2-4} \cmidrule(lr){5-7}
Model & BLEU-4 $\uparrow$ & ROUGE-L $\uparrow$ & CIDEr-D $\uparrow$ & BLEU-4 $\uparrow$ & ROUGE-L $\uparrow$ & CIDEr-D $\uparrow$ \\
\toprule
Most Frequent & 20.0 & 31.0 & \textbf{42.0} & 20.0 & 30.0 & \textbf{43.0} \\
Text-Only     & 14.0 & 36.0 & 5.0 & 14.0 & 36.0 & 7.0 \\
Neural Naturalist & 24.0 & 46.0 & 28.0 & 22.0 & 43.0 & 25.0 \\
CNN+LSTM & 25.1 & 43.4 & 10.2 & 24.9 & 43.2 & 9.9  \\
\midrule 
\textsc{L2C} [B2W] & 31.9 & 45.7 & 15.2 & 31.3 & 45.3 & 15.1 \\
\textsc{L2C} [CUB+B2W] & \textbf{32.3} & \textbf{46.2} & 16.4 & \textbf{31.8} & \textbf{45.6} & 16.3 \\
\midrule
Human & 26.0 & 47.0 & 39.0 & 27.0 & 47.0 & 42.0 \\
\bottomrule
\end{tabular}
\caption{Results for visual comparison on the Birds-to-Words dataset~\citep{forbes2019neural}. \textit{Most Frequent} produces only the most observed description in the dataset: ``the two animals appear to be exactly the same". \textit{Text-Only} samples captions from the training data according to their empirical distribution. \textit{Neural Naturalist} is a transformer model in ~\citet{forbes2019neural}. \textit{CNN+LSTM} is a commonly-used CNN encoder and LSTM decoder model.
}
\label{tab:main}
\end{table*}

\paragraph{Implementation Details}
We use Adam as the optimizer with an initial learning rate set to 1e-4. The pooling module to generate $K$ classes is composed of two convolutional layers and batch normalization, with kernel sizes 3 and 1 respectively. We set $K$ to 9 and $\lambda$ to 1. The dimension of graph representations is 512. The hidden size of the decoder is also 512. The batch sizes of B2W and CUB are 16 and 128. Following the advice from ~\citep{forbes2019neural}, we report the results using models with the highest ROUGE-L on the validation set, since it could correlate better with high-quality outputs for this task.

\subsection{Automatic Evaluation}
As shown in \cref{tab:main}, first, L2C[B2W] (training with visual comparison task only) outperforms baseline methods on BLEU-4 and ROUGE-L. Previous approaches and architectures failed to bring superior results by directly modeling the visual relationship on ResNet features.
Second, joint learning with a single-image caption L2C[B2W+CUB] can help improve the ability of semantic understanding, thus, the overall performance of the model.
Finally, our method also has a smaller gap between validation and test set compared to \textit{neural naturalist}, indicating its potential capability to generalize for unseen samples.

\begin{table}
\small
\centering
\begin{tabular}{c c|c|c}
% \toprule
%   & \multicolumn{3}{c}{L2C vs CNN+LSTM}\\
\toprule
Choice (\%) & L2C & CNN+LSTM & Tie \\
\midrule
Score & \textbf{50.8} & 39.4 & 9.8 \\
\bottomrule
\end{tabular}
\caption{Human evaluation results. We present workers with two generations by L2C and CNN+LSTM for each image pair and let them choose the better one.
}
\label{tab:human}
\end{table}

\subsection{Human Evaluation}
To fully evaluate our model, we conduct a pairwise human evaluation on Amazon Mechanical Turk with 100 image pairs randomly sampled from the test set, each sample was assigned to 5 workers to eliminate human variance. Following~\citet{wang2018arel}, for each image pair, workers are presented with two paragraphs from different models and asked to choose the better one based on text quality\footnote{We instruct the annotators to consider two perspectives, relevance (the text describes the context of two images) and expressiveness (grammatically and semantically correct).}. As shown in  \cref{tab:human}, \textsc{L2C} outperforms \textsc{CNN+LSTM}, which is consistent with automatic metrics.

\subsection{Ablation Studies} 

% \noindent\textbf{Effect of Individual Components~~}
\paragraph{Effect of Individual Components}
We perform ablation studies to show the effectiveness of semantic pooling, total variance loss, and graph reasoning, as shown in \cref{tab:ablation}.
First, without semantic pooling, the model degrades to average pooling, and results show that semantic pooling can better preserve the spatial relations for the visual representations. 
Moreover, the total variation loss can further boost the performance by injecting the prior local smoothness.
Finally, the results without GCN are lower than the full L2C model, indicating graph convolutions can efficiently modeling relations among visual regions.

\begin{table}[t]
\small
\centering
\setlength{\tabcolsep}{2pt}
% \resizebox{\linewidth}{!}{
\begin{tabular}{l rrr}
\toprule
  & \multicolumn{3}{c}{\textbf{Validation}}\\
\cmidrule(lr){2-4}
Model  & BLEU-4 $\uparrow$ & ROUGE-L $\uparrow$ & CIDEr-D $\uparrow$ \\
\toprule
L2C  & \textbf{31.9} & \textbf{45.7}  & \textbf{15.2} \\
\midrule 
$-$ Semantic Pooling & 24.5 & 43.2 & 7.2 \\
$-$ TV Loss & 29.3 & 44.8 & 13.6 \\
$-$ GCN  & 30.2 & 43.5 & 10.7 \\
\bottomrule
\end{tabular}
% }
\caption{Ablation study on the B2W dataset. We individually remove Semantic Pooling, total variation (TV) loss, and GCN to test their effects.
}
\label{tab:ablation}
\end{table}

\begin{figure}[t]
  \centering
  \includegraphics[width=.8\linewidth]{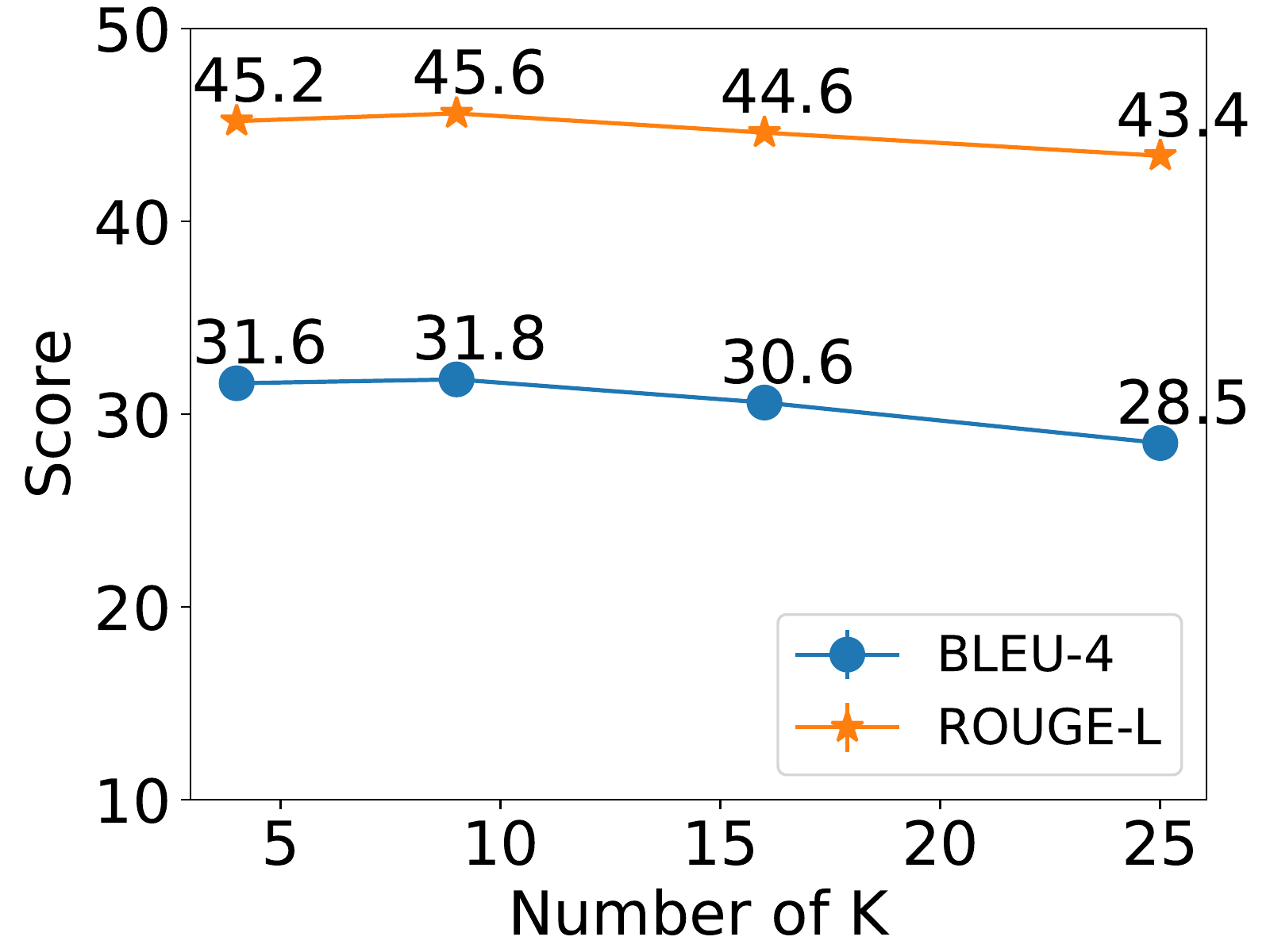}
  \caption{Sensitivity test on number of K chosen.}
  \label{fig:robust}
\end{figure}

% \noindent\textbf{Sensitivity Test~~}
\paragraph{Sensitivity Test}
We analyze model performance under a varying number of $K$ ($K$ is the number of classes for confidence map $\mathcal{C}$), as shown in \cref{fig:robust}. Empirically, we found the results are comparable when $K$ is small.

\section{Conclusion}
In this paper, we present a learning-to-compare framework for generating visual comparisons. 
Our segmentation encoder with semantic pooling and graph reasoning could construct structured image representations. 
We also show that learning to describe visual differences benefits from understanding the semantics of each image.

\section*{Acknowledgments}
The research was partly sponsored by the U.S. Army Research Office and was accomplished under Contract Number W911NF19-D-0001 for the Institute for Collaborative Biotechnologies. The views and conclusions contained in this document are those of the authors and should not be interpreted as representing the official policies, either expressed or implied, of the U.S. Government. The U.S. Government is authorized to reproduce and distribute reprints for Government purposes notwithstanding any copyright notation herein.

\bibliography{eacl2021}
\bibliographystyle{acl_natbib}

\end{document}